\pdfoutput=1

\documentclass[11pt]{article}

\usepackage{ACL2023}

\usepackage{times}
\usepackage{latexsym}

\usepackage[T1]{fontenc}

\usepackage[utf8]{inputenc}

\usepackage{microtype}

\usepackage{inconsolata}

\usepackage{tabularx}
\usepackage{booktabs}
\usepackage{graphicx}
\usepackage{multirow}
\usepackage{enumitem}
\setlist{nosep}
\usepackage{soul}

\newcommand{\rc}[1]{#1}

\title{Text-to-SQL Error Correction with Language Models of Code}

\author{Ziru Chen$^1$, Shijie Chen$^1$, Michael White$^1$, Raymond Mooney$^2$ \\
  \textbf{Ali Payani$^3$, Jayanth Srinivasa$^3$, Yu Su$^1$, Huan Sun$^1$} \\
  $^1$The Ohio State University \\
  $^2$The University of Texas at Austin \quad $^3$Cisco Research\\
  \texttt{\{chen.8336, chen.10216, white.1240, su.809, sun.397\}@osu.edu} \\
  \texttt{mooney@cs.utexas.edu} \quad \texttt{\{apayani, jasriniv\}@cisco} \\
}

\begin{document}
\maketitle
\begin{abstract}
Despite recent progress in text-to-SQL parsing, current semantic parsers are still not accurate enough for practical use. 
In this paper, we investigate how to build automatic text-to-SQL error correction models.
Noticing that token-level edits are out of context and sometimes ambiguous, we propose building clause-level edit models instead. 
Besides, while most language models of code are not specifically pre-trained for SQL, they know common data structures and their operations in programming languages such as Python. 
Thus, we propose a novel representation for SQL queries and their edits that adheres more closely to the pre-training corpora of language models of code.
Our error correction model improves the exact set match accuracy of different parsers by $2.4$--$6.5$ and obtains up to $4.3$ point absolute improvement over two strong baselines.\footnote{Our code and data are available at \url{https://github.com/OSU-NLP-Group/Auto-SQL-Correction}.} 
\end{abstract}

\section{Introduction}
\label{intro}
Text-to-SQL parsing is a classic semantic parsing task that finds wide applications~\citep{geoquery, restaurant}.
Since the release of Spider \citep{yu-etal-2018-spider}, a cross-database text-to-SQL benchmark, many semantic parsers with decent performance have been developed \citep{lin-etal-2020-bridging, wang-etal-2020-rat, deng-etal-2021-structure, rubin-berant-2021-smbop, scholak-etal-2021-picard}.
Nonetheless, state-of-the-art semantic parsers are still not accurate enough. 
As a result, their users need to constantly correct wrongly predicted SQL queries, which can be as time-consuming and error-prone as writing a SQL query from scratch \citep{bugfixhard1, bugfixhard2}.
Therefore, in this paper, we study the problem of automatic text-to-SQL error correction to better assist users in querying complex databases.

We first highlight that it is essential to factor in the \textit{compositional substructures} within SQL queries, such as abstract syntax trees \citep{yin-neubig-2017-syntactic, guo-etal-2022-unixcoder} and data-flow graphs \citep{guo2021graphcodebert}, instead of treating code snippets as string sequences. 
Compared to individual tokens, substructures (e.g. SQL clauses) include more context of the entire program and are more semantically meaningful. 
Consequently, edit patterns of such substructures are more intuitive for humans to understand and easier for language models to learn. 
Moreover, while the pre-training corpora for language models of code, such as CodeT5 \citep{wang-etal-2021-codet5}, do not include many SQL queries based on their documentation, they naturally contain \textit{abundant examples of common data structures} like dictionaries. 
Therefore, we hypothesize that transforming unfamiliar SQL queries into familiar data structures can help language models of code better perform structural editing of SQL queries.

\begin{table*}
\small
\centering
\begin{tabularx}{\textwidth}{cXcX}
\toprule
\multicolumn{2}{c}{\textbf{Query Representation}} & \multicolumn{2}{c}{\textbf{Edit Representation}} \\
\cmidrule(lr){1-2}\cmidrule(lr){3-4}
\multirow{2}{*}{SQL} & \multirow{2}{\hsize}{\textcolor{gray}{select tweets.text \\ from tweets \\ order by \textcolor{red}{tweets.text}}} & Token-Level & \textcolor{gray}{<ReplaceOld> \textcolor{red}{tweets.text} <ReplaceNew> \textcolor{red}{tweets.createdate} <ReplaceEnd>} \\
 & & Clause-Level & \textcolor{gray}{<ReplaceOld> order by \textcolor{red}{tweets.text} <ReplaceNew> order by \textcolor{red}{tweets.createdate} <ReplaceEnd>} \\
\cmidrule(lr){1-2}\cmidrule(lr){3-4}
\multirow{2}{*}{PyDict} & \multirow{2}{\hsize}{\textcolor{gray}{sql = \{\\ \quad "select": "select tweets.text", \\ \quad "from": "from tweets", \\ \quad "orderBy": "order by \textcolor{red}{tweets.text}"\\ \}}} & Clause-Level & \textcolor{gray}{<ReplaceOld> "orderBy": "order by \textcolor{red}{tweets.text}" <ReplaceNew> "orderBy": "order by \textcolor{red}{tweets.createdate}" <ReplaceEnd>} \\
 &  & Program & \textcolor{gray}{sql["orderBy"] = "order by  \textcolor{red}{tweets.createdate}"} \newline \\
\bottomrule
\end{tabularx}
\caption{\label{tab:rep}
Example representations for a wrong SQL query and the \textit{Replace} edit action. The corresponding natural language utterance is ``List the text of all tweets in the order of date.''
For token-level and clause-level representations, we format them as ``\texttt{<ReplaceOld>} Span of wrong tokens/clauses \texttt{<ReplaceNew>} Span of correct tokens/clauses \texttt{<ReplaceEnd>}'', where \texttt{<ReplaceOld>}, \texttt{<ReplaceNew>}, and \texttt{<ReplaceEnd>} are special tokens.
}
\vspace{-11pt}
\end{table*}

Based on these observations, we develop our error correction model and make two contributions.
First, we propose considering SQL clauses instead of tokens as basic semantic units for editing.
Using a context-free grammar, we can decompose a SQL query and identify its clauses by traversing its abstract syntax tree.
Second, we propose a new representation of SQL queries and their edits that adheres more closely to common code pre-training corpora, including CodeSearchNet \citep{codesearchnet}, and makes the structures of a SQL query more explicit.
With a decomposed SQL query, we pair each clause with its SQL keyword and represent the entire query as a Python dictionary. 
Then, we format edits on a wrong SQL query as a program that modifies data of the query's corresponding dictionary. 
Unlike token-level edits in existing work~\cite{ZhangETAL22CoditT5}, such dictionary operations define all edits unambiguously and can be directly executed with a Python interpreter. 

Through comprehensive experiments with different representations, we show that:
(1) our proposed representation has the lowest zero-shot perplexity with CodeT5;
(2) simply changing token-level edits to clause-level edits can effectively improve the performance of our models; 
and (3) our method improves the exact set match accuracy of different parsers by $2.4$--$6.5$ and obtains up to $4.3$ point absolute improvement over two strong baselines.

\section{Text-to-SQL Error Correction}
\label{task_formulation}
Given a natural language utterance $\mathbf{u}$, a database schema $\mathbf{s}$, and a wrong SQL query $\mathbf{q_{-}}$ produced by an existing parser, our goal is to develop an error correction model that predicts a sequence of edit actions $\mathbf{e}$ and the correct query $\mathbf{q_{+}}$. Following previous work \citep{ZhangETAL22CoditT5}, we formulate our task as sequence-to-sequence generation:
\begin{equation}
    \label{eq:generation}
    P(\mathbf{y} | \mathbf{x}) = \Pi_{t=1}^{T} P(\mathbf{y}_{t} | \mathbf{x}, \mathbf{y}_{1:t-1})
\end{equation}
where $\mathbf{x} = [\mathbf{u}; \mathbf{s}; \mathbf{q_{-}}]$ is the concatenation of the given inputs and $\mathbf{y} = [\mathbf{e}; \mathbf{q_{+}}]$ is the concatenation of all edit actions and the resulting correct query. 
In this section, we study different representations of SQL queries (Section \ref{queryRep}) and edits (Section \ref{editRep}) to better leverage language models of code. 

\subsection{Query Representation} 
\label{queryRep}
We consider two representations for a predicted query: (1) the original SQL format and (2) our proposed PyDict (Python Dictionary) representation. 
To prepare for editing, we disambiguate each SQL query following \citet{rubin-berant-2021-smbop}, including lower-casing non-value tokens, resolving table references, and formatting punctuation. 
This preprocessing normalizes SQL queries predicted by different base parsers and the gold annotations into the same format. 
To build our PyDict representation, we parse a SQL query into its abstract syntax tree (AST) with Spider's context-free grammar. 
We use depth-first search to traverse through the AST, find any nested substructures, and construct the dictionary representation bottom-up. 
Table \ref{tab:rep} shows the ``SQL'' and ``PyDict'' representations of a SQL query (more details in Appendix \ref{appendix:decomposition}).

\subsection{Edit Representation} 
\label{editRep}
We first follow \citet{ZhangETAL22CoditT5} to use token-level edit representation with special tokens (Table~\ref{tab:rep}), which have unique entries in the tokenizer and the model's embedding layer to describe \textit{Replace, Insert}, and \textit{Delete} edit actions (more examples in Appendix \ref{appendix:examples}).
However, we realize this representation can sometimes be ambiguous. 
As shown in Table \ref{tab:rep}, the span ``tweets.text'' appears twice in the SQL query.
This repetition would confuse the error correction model with which span to replace when generating the corrected query. 
Also, the ambiguity makes it difficult to implement rules and directly carry out the edit actions on the wrong query.
Hence, we change the token-level edit representation to clause-level, which includes more context of the query to make different edits more distinguishable. 
In our experiments (Section \ref{results}), we demonstrate that this simple modification is already effective.
Our program representation further improves the performance because it is more similar to the code pre-training corpora and eliminates the need to learn special tokens' representations.

\section{Experimental Setup}
\label{setup}
\subsection{Data Synthesis for SQL Error Correction}
\label{dataSyn}
To train a text-to-SQL error correction model, we need to collect a set of wrong SQL parses that reflects a realistic distribution of errors (Section \ref{error}) as our training data.
We synthesize this dataset by performing 5-fold cross-validation on each parser, which approximates the actual evaluation setting.

Following the evaluation setup in \citet{yu-etal-2018-spider}, we split Spider's training set into five roughly equal subsets by different databases. 
For each cross-validation fold, we train a text-to-SQL parser (Section \ref{models}) on four subsets and evaluate it on the remaining one.
At inference time, we perform beam search with size 20 for each example and collect grammatical and executable parses in the beam.\footnote{Due to SmBoP's bottom-up decoding, we keep its original beam size and collect the top-20 unique beam predictions.} 
If a SQL parse is not an exact set match or execution match to the gold annotation, we label it wrong and include it in our training set for error correction.
Having synthesized our training dataset, we randomly sample 8 databases and their associated questions to construct a held-out development set. 
For development set examples, we only keep incorrect SQL parses with the highest beam confidence. 
For our error correction test set, we train each parser on the \textit{full} Spider training set and evaluate it on the original Spider's development set without modifications. 
We similarly keep SQL parses with exact match or execution match errors.
Table \ref{tab:data} summarizes the statistics of our data.

\begin{table}[t]
\small
\centering
\begin{tabular}{lccc}
\toprule
 & \textbf{CodeT5} & \textbf{BRIDGEv2} & \textbf{SmBoP}\\
\midrule
\# of Train & 47,020 & 24,776 & 20,083 \\
\# of Dev & 448 & 448 & 448 \\
\# of Test & 430 & 392 & 310 \\ 
\midrule
Avg. Train Edits & 2.34 & 3.11 & 2.72 \\
Avg. Dev Edits & 2.70 & 3.29 & 3.31 \\
Avg. Test Edits & 1.84 & 1.51 & 1.47 \\
\bottomrule
\end{tabular}
\caption{Summary of data statistics. 
}
\vspace{-11pt}
\label{tab:data}
\end{table}

\subsection{Models}
\label{models}

\paragraph{Text-to-SQL base parsers.} 
We choose three text-to-SQL parsers with different decoding strategies and levels of performance (Table \ref{tab:main}).
We elaborate on our selection criteria in Appendix \ref{appendix:parser_select}.
\begin{itemize}
    \item \textbf{CodeT5} \citep{wang-etal-2021-codet5}: We fine-tune \texttt{CodeT5-base} following \citet{unifiedskg}. This parser represents those using beam search decoding and having a lower accuracy. 
    \item \textbf{BRIDGEv2} \citep{lin-etal-2020-bridging}: A representative parser with constrained decoding and achieving a medium-level accuracy.
    \item \textbf{SmBoP} \citep{rubin-berant-2021-smbop}: A representative parser with bottom-up decoding and achieving higher accuracy.
\end{itemize}

\paragraph{Error correction models.} We use two language models of code in all our experiments: 
\begin{itemize}
    \item \textbf{CoditT5} \citep{ZhangETAL22CoditT5}: A language model pre-trained for code editing tasks by injecting noises to code snippets in CodeSearchNet \citep{codesearchnet} and then denoising with token-level edit representations. 
    \item \textbf{CodeT5} \citep{wang-etal-2021-codet5}: A language model pre-trained for general code understanding and generation with four different pre-training objectives. 
\end{itemize}
We compare the existing SQL+Token-Level representation with our proposed ones: SQL+Clause-Level, PyDict+Clause-Level, and PyDict+Program on CodeT5 and the first three on CoditT5.\footnote{We did not use CoditT5 for PyDict+Program because it was pre-trained on token-level edit representations. Its decoder may be specialized in generating edits instead of programs.} 
Implementation details are in Appendix \ref{appendix:implementation}.

\begin{table*}
\small
\centering
\begin{tabular}{lcccccccc}
\toprule
\multirow{2}{*}{\textbf{Models}} & \multirow{2}{*}{\textbf{Query}} & \multirow{2}{*}{\textbf{Edit}} & \multicolumn{2}{c}{\textbf{CodeT5}} & \multicolumn{2}{c}{\textbf{BRIDGEv2}} & \multicolumn{2}{c}{\textbf{SmBoP}} \\
\cmidrule(lr){4-5}\cmidrule(lr){6-7}\cmidrule(lr){8-9}
 & & & EM & EX & EM & EX & EM & EX\\
\midrule
No Edit & N/A & N/A & 62.7 (-) & 63.6 (-) & 70.1 (-) & 68.2 (-) & 74.6 (-) & 75.3 (-) \\
\midrule
\multirow{3}{*}{CoditT5} & SQL & Token-Level & 64.3 (0.1) & 64.4 (0.2) & 65.4 (0.5) & 66.6 (0.3) & 74.2 (0.4) & 75.3 (0.1) \\
 & SQL & Clause-Level & 67.0 (0.4) & 65.4 (0.5) & 71.3 (0.5) & 70.9 (0.2) & 76.3 (0.0) & 77.2 (0.3) \\
 & PyDict & Clause-Level & 67.1 (0.2) & 66.5 (0.4) & 70.6 (0.8) & 70.8 (0.6) & 76.3 (0.3) & 77.0 (0.3) \\
\midrule
\multirow{3}{*}{CodeT5} & SQL & Token-Level & 66.7 (0.9) & 65.9 (0.5) & 68.2 (0.4) & 69.4 (0.8) & 75.6 (0.4) & 76.5 (0.6) \\
 & SQL & Clause-Level & 68.3 (0.3) & \underline{68.2}$^+$(0.6) & 71.8$^+$(0.4) & 72.5$^+$(0.2) & 76.7 (0.6) & 77.4 (0.3)  \\
 & PyDict & Clause-Level & 66.6 (0.8) & 67.1 (0.8) & 72.0$^+$(0.3) & 72.4$^+$(0.2) & \underline{77.3} (0.6) & \underline{77.8} (0.2) \\
\midrule
CodeT5$^*$ & \multirow{2}{*}{PyDict} & \multirow{2}{*}{Program} & \textbf{69.2}$^+$(0.4) & \textbf{68.4}$^+$(0.2) & \textbf{72.5}$^+$(0.4) & \textbf{73.1}$^+$(0.2) & \underline{77.3} (0.4) & 77.6 (0.6) \\
CodeT5 &  &  & \underline{69.0}$^+$(0.2) & \underline{68.2}$^+$(0.1) & \textbf{72.5}$^+$(0.3) & \underline{73.0}$^+$(0.6) & \textbf{78.0}$^+$(0.3) & \textbf{78.5}$^+$(0.3) \\
\bottomrule
\end{tabular}
\caption{\label{tab:main}
Exact Set Match (EM) and Execution Match (EX) accuracy on Spider development set. The \textbf{best performances} are in bold and the \underline{second bests} are underlined. Results with $^+$ are statistically significant (McNemar's; $p < 0.05$) compared to CodeT5-SQL+Token-Level (Appendix \ref{appendix:statTest}). Otherwise, the results are not statistically significant. $^*$We fine-tune the model to generate edit programs only (without resulting queries) and use Python interpreter to execute the edit actions.
}
\vspace{-11pt}
\end{table*}

\subsection{Evaluation}
\label{eval}
\rc{We use the increase in Exact Set Match (EM) and Execution Match (EX) accuracy on our error correction test set to measure each model's performance.
Because CoditT5’s experiments assume the input program has at least one error, we keep this assumption for fair comparisons. 
To construct a test set satisfying this assumption, we have to compare parser-generated SQL queries with gold annotations (Section \ref{dataSyn}). 
Thus, we use the Spider development set as our test set and split the Spider training set to build a held-out development set (Table \ref{tab:data}) to select model checkpoints during training.
We also include results on our held-out development set in the appendix (Table \ref{tab:dev}).} 

\section{Results and Analysis}
\subsection{Main Results}
\label{results}
We summarize our main results in this section.
To ensure robustness, we repeat all experiments with 3 different random seeds and report the average performances with standard deviations. 
Our model can also be used in an interactive framework that allows users to select edit actions from the top-$k$ beam candidates.
We include more experiments with simulated user interactions in Appendix \ref{appendix:otherResults}.

\begin{figure}[t]
  \centering
  \includegraphics[width=0.9\columnwidth]{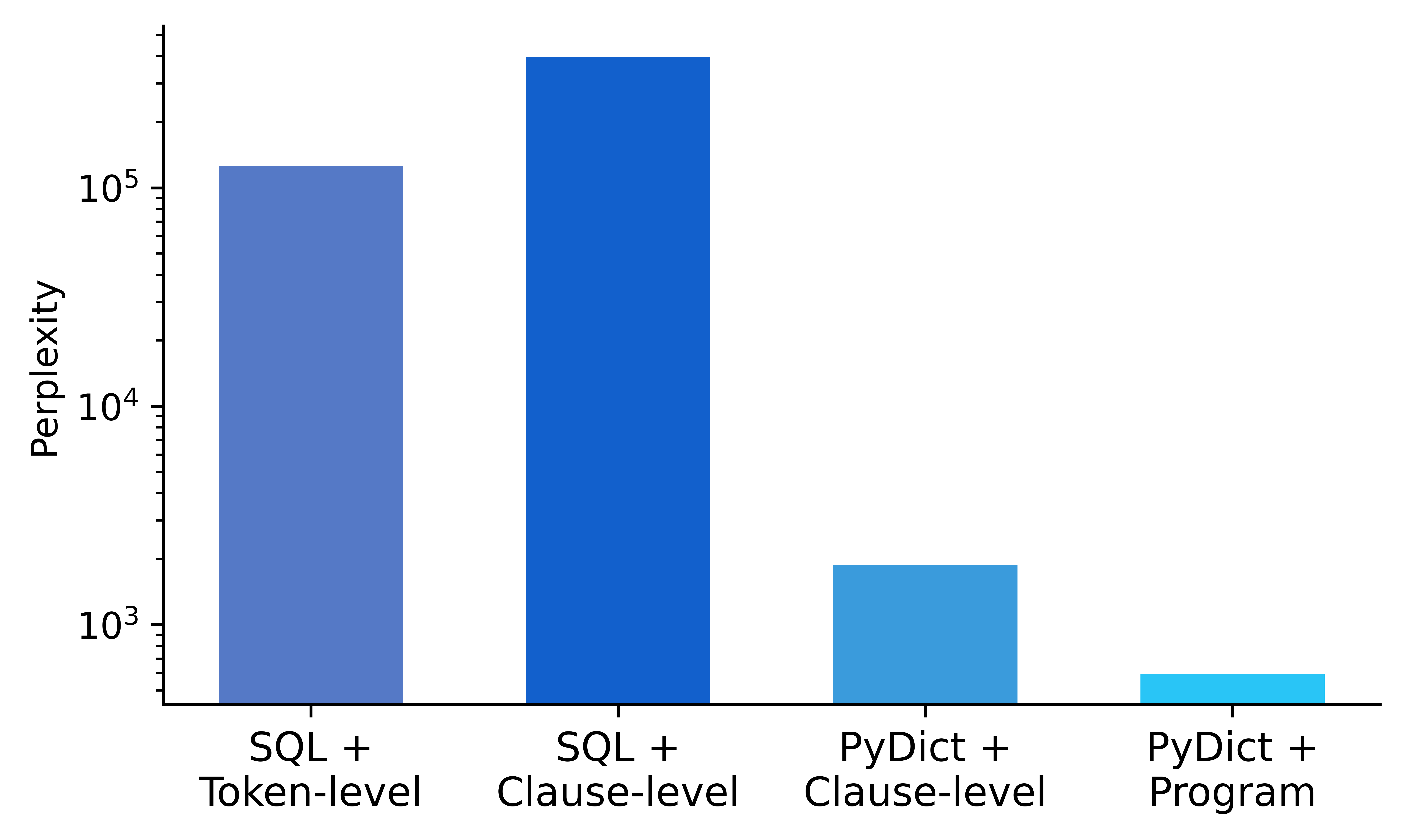}
  \caption{\label{fig:ppl}
    CodeT5's zero-shot perplexity (in log scale) of all four representations on our synthesized SQL error development set. 
  }
  \vspace{-12pt}
\end{figure}

\paragraph{Our representation's perplexity is the smallest.}
We validate that our PyDict+Program representation adheres more closely to the code pre-training corpora by measuring its zero-shot perplexity on CodeT5 using our development set (Section \ref{dataSyn}). 
As shown in Figure \ref{fig:ppl}, by representing data in PyDict, we can reduce the perplexity of CodeT5 by 2 orders of magnitude. 
After augmenting it with our program representation, we further reduce the zero-shot perplexity of CodeT5 to only $5.96 \times 10^2$, 3 orders of magnitude less than the SQL+Token-Level representation ($1.26 \times 10^5$). 

\paragraph{Clause-level editing is more effective, especially when represented in PyDict+Program.} 
\rc{Since CodeT5 consistently outperforms CoditT5 with the same representations, we focus on comparisons among CodeT5 variations.
As shown in Table \ref{tab:main}, compared to CodeT5-SQL+Token-Level, only CodeT5-PyDict+Program achieves statistically significant improvement on all three parsers, while clause-level models fail McNemar’s significance test for some parsers. 
More concretely, it achieves up to 4.3 point more absolute improvement on EM accuracy ($68.2 \rightarrow 72.5$; BRIDGEv2) and 3.7 point more absolute improvement on EX accuracy ($69.4 \rightarrow 73.1$; BRIDGEv2).
Overall, CodeT5-PyDict+Program can boost the parsers' EM accuracy by 2.4--6.5. 
Thus, both clause-level editing and PyDict+Program representation can better take advantage of language models of code.}


\begin{table*}[t]
\small
\centering
\begin{tabular}{lccccccccc}
\toprule
\multirow{2}{*}{\textbf{Error Category}} & \multicolumn{3}{c}{\textbf{CodeT5}} & \multicolumn{3}{c}{\textbf{BRIDGEv2}} & \multicolumn{3}{c}{\textbf{SmBoP}} \\
\cmidrule(lr){2-4}\cmidrule(lr){5-7}\cmidrule(lr){8-10}
 & Resolved & Unresolved & All & Resolved & Unresolved & All & Resolved & Unresolved & All\\
\midrule
Database Grounding & 15 & 51 & 66 & 14 & 48 & 62 & 16 & 38 & 54 \\
Incorrect Structure & 2 & 15 & 17 & 2 & 12 & 14 & 3 & 23 & 26 \\
Syntax \& Grammar & 0 & 0 & 0 & 0 & 0 & 0 & 1 & 4 & 5 \\
False Negative & 0 & 9 & 9 & 0 & 6 & 6 & 0 & 8 & 8\\
Other & 1 & 7 & 8 & 2 & 16 & 18 & 1 & 6 & 7 \\
\bottomrule
\end{tabular}
\caption{\label{tab:error} Analysis of 100 sample errors made by each text-to-SQL parser. We group the errors into 5 categories and examine if our CodeT5-PyDict+Program model resolves them.}
\vspace{-11pt}
\end{table*}

\subsection{Error Analysis}
\label{error}
\rc{Additionally, we conduct an error analysis (Table \ref{tab:error}) by sampling 100 wrong parses from all three parsers and classifying them into five categories:}
\begin{itemize}
    \item \textit{Database Grounding}: \rc{A generated SQL query has the correct structure, but some table/column names or entity values are wrong.}
    \item \textit{Incorrect Structure}: \rc{A generated SQL query has missing, wrong, or redundant structures.}
    \item \textit{Syntax \& Grammar}: \rc{A generated SQL query violates the programming language’s syntax.}   
    \item \textit{False Negative}: \rc{A generated SQL query is semantically correct but not captured by evaluation metrics, or the gold annotation is wrong.}
    \item \textit{Other}: \rc{All other errors, such as wrong aggregation functions, besides the above categories.}
\end{itemize}
\rc{Since the error distributions for each parser are similar, as an example, we discuss our findings based on the strongest parser, SmBoP:}

\paragraph{Database grounding is the major type of error.}
\rc{Among the 100 samples from SmBoP, we find that 54 of them have database grounding errors.
Particularly, SmBoP predicts wrong table/column names in 34 parses, inaccurate entity values in 9 parses, and incorrect \texttt{JOIN} relations in 11 parses.
Our CodeT5-PyDict+Program model can successfully fix 16 of the 54 erroneous parses, including 10 parses with wrong table/column names, 4 parses with inaccurate entity values, and 2 parses with incorrect \texttt{JOIN} relations.
We hypothesize that database grounding is also a major category of errors in our synthesized training set, so our model has learned to resolve similar errors.
Nevertheless, it still cannot correct the remaining 38 SQL parses.
We notice that our current representation for database schema is missing critical information, such as column data types and foreign key relations, for our error correction model to fix database grounding errors.
Following our PyDict representation for SQL, we suggest designing a code representation for database schema that includes such information to tackle this issue in future work.}

\paragraph{Structural errors are hard to edit automatically.}

\rc{Besides database grounding, 26 of SmBoP's errors belong to another category, incorrect structure.
These 26 samples contain 7 parses with incorrect SQL clauses and 19 parses with incorrect subqueries, but our CodeT5-PyDict+Program model only resolves 1 and 2 of them, respectively.
We find that correcting such errors usually involves multiple edit steps, which motivates us to incorporate our model into an interactive framework in future work.
As our experiments with simulated user interaction (Appendix \ref{tab:interaction}) show, when our model interacts with the simulated user to correct one clause at a time, it is able to fully correct more SQL parses.
Thus, we deem interactive correction would maximize our model's utility in practice.}

\section{Related Work}
\rc{Since the release of CodeBERT \citep{feng-etal-2020-codebert}, many language models of code have emerged for program understanding and generation \citep{ahmad-etal-2021-unified, codex, guo2021graphcodebert, wang-etal-2021-codet5, guo-etal-2022-unixcoder, incoder, codegen}.
In addition to program-related tasks, recent work shows they also excel at processing natural language structures.
Using code as meaning representations (MRs), we can leverage language models of code in various tasks, such as commonsense reasoning \citep{madaan-etal-2022-language}, action planning \citep{singh2022progprompt}, and event extraction \citep{wang2022code4struct}.
In fact, how to design MRs to reduce model learning difficulty is a salient research question in semantic parsing \citep{guo-etal-2019-towards, gan-etal-2021-natural-sql, nie-etal-2022-graphq}.}

\rc{Our work demonstrates that program-related tasks themselves can also benefit from code-based MRs.
Specifically, we apply such MRs to SQL error correction, a variant of automatic program repair tasks \citep{10.1109/ICSE.2019.00021, panthaplackel-etal-2022-using, ZhangETAL22CoditT5}.
Although SQL is a code-based MR, it is much harder for models to learn compared to other MRs, such as FunQL and lambda calculus \citep{li-etal-2022-exploring-secrets}.
Consequently, without many SQL queries in their pre-training corpora, language models of code can underperform state-of-the-art text-to-SQL parsers.
By converting SQL queries into Python dictionaries, we can explicitly represent their compositional substructures and define edit actions as programs, which reduces the learning difficulty for language models of code and yields better performance.}

\section{Conclusion and Future Work}
This paper presents a study on developing a text-to-SQL error correction model with clause-level edits and different representations. 
Our comprehensive experiments demonstrate that \textit{clauses are better semantic units than tokens} for editing SQL queries and \textit{mimicking patterns in code pre-training corpora} helps better leverage language models of code.
As a future direction, we plan to incorporate our model into interactive semantic parsing frameworks \citep{li-etal-2020-mean, yao-etal-2019-model, yao-etal-2020-imitation, zeng-etal-2020-photon} by suggesting possible edits to users once a wrong parse is identified. 
In this way, users would more efficiently correct parse errors and get better assistance.
\rc{We also plan to experiment with other language models of code \citep{incoder, codegen} and text-to-SQL datasets \citep{geoquery, gan-etal-2021-towards} to verify the generalizability of our method.}

\section*{Limitations}
\paragraph{Actual applications of our model. }
Our work assumes that input SQL queries to our model are always wrong. 
This assumption is more feasible in an interactive semantic parsing framework, where the users are expected to decide whether a SQL parse, accompanied by its natural language explanations \citep{elgohary-etal-2020-speak, elgohary-etal-2021-nl, diy, mo-etal-2022-towards}, has errors or not. 
Alternatively, to remove this assumption, it would be interesting for future work to study the performance of our error correction model in combination with an automatic error detection model \citep{chen2023error}. 

\paragraph{Experiments with more language models of code.}
We have only experimented with two language models of code, CoditT5 and CodeT5, both using T5-base \citep{t5} as their underlying model architecture. It would be interesting to test how our conclusions generalize to other language models of code in the future. Based on the strong capabilities of large language models of code, such as Codex \citep{codex}, InCoder \citep{incoder}, and CodeGen \citep{codegen}, we believe that these models can better exploit their knowledge about data structures and their operations in Python. These models may perform even better on Text-to-SQL error correction with our proposed representations.


\section*{Acknowledgements}
\rc{We would like to thank the anonymous reviewers and colleagues from the OSU NLP group for their thoughtful comments. 
This research was supported in part by a sponsored award from Cisco Research, NSF IIS-1815674, NSF CAREER \#1942980, NSF OAC-2112606, and Ohio Supercomputer Center \citep{OhioSupercomputerCenter1987}. 
The views and conclusions contained herein are those of the authors and should not be interpreted as representing the official policies, either expressed or implied, of the U.S. government.
The U.S. Government is authorized to reproduce and distribute reprints for Government purposes notwithstanding any copyright notice herein.
Ziru is also supported by The Ohio State University Graduate School through University Fellowship.}

\bibliography{anthology,custom}
\bibliographystyle{acl_natbib}

\appendix
\setcounter{table}{0}
\renewcommand\thetable{\Alph{section}.\arabic{table}}
\setcounter{figure}{0}
\renewcommand\thefigure{\Alph{section}.\arabic{figure}}
\section*{Appendices}
We provide more details omitted in the main text as follows:
\begin{itemize}
    \item Appendix \ref{appendix:decomposition}: SQL PyDict Representation
    \item Appendix \ref{appendix:parser_select}: Text-to-SQL Parser Selection
    \item Appendix \ref{appendix:implementation}: Implementation Details
    \item Appendix \ref{appendix:statTest}: Statistical Significance Test
    \item Appendix \ref{appendix:otherResults}: Additional Results
    \item Appendix \ref{appendix:examples}: More Representation Examples
\end{itemize}

\section{SQL PyDict Representation}
\label{appendix:decomposition}
We implement the transformation from any SQL query to our PyDict representation in three steps (Section \ref{queryRep}).
First, we use context-free grammar to parse a SQL query and obtain its abstract syntax tree (AST).
The AST naturally contains a SQL decomposition where each clause has its unique subtree. 
In addition, if a clause contains a nested query, it would be represented as another independent subtree, which is a child of the root node in the clause's AST subtree. 
With these substructures explicitly represented, we use depth-first search to traverse through the AST to build our PyDict representation bottom-up.
In other words, if a clause contains a subquery, we process the subquery tree as an independent SQL AST and build a dictionary for it.
Then, we combine it with other substructures of the clause with different dictionary keys.
For example, in Table \ref{tab:rep2}, we first build the dictionary for ``subquery0'' and assign this identifier as the key. 
In the main ``clause,'' we replace the subquery's corresponding span with this identifier.
Finally, we use another dictionary to wrap the main ``clause'' and ``subquery0'' together as the final representation of the ``where'' clause.
We repeat this procedure for each clause to incrementally add (key, value) pairs to the dictionary and ``store'' it to the variable \texttt{sql}, which we refer to in program edit representations. 

\section{Text-to-SQL Parser Selection}
\label{appendix:parser_select}
\rc{
We choose existing text-to-SQL parsers in our experiments according to two principles: the parsers predict database entity values, and they cover different decoding strategies, including grammar-based (BRIDGEv2), bottom-up (SmBop), and token-based (CodeT5). 
We did not include parsers using top-down decoders because they usually cannot predict entity values in conditional statements, such as RAT-SQL \citep{wang-etal-2020-rat}.
Instead, we include BRIDGEv2 because its decoding method mimics the left-to-right CFG derivation of a program, and it uses SQL syntax-based constraints to prevent grammatical errors.
In recent work, such decoders, also used in PICARD \citep{scholak-etal-2021-picard}, are more popular than top-down decoders.}

\section{Implementation Details}
\label{appendix:implementation}
Our models (Section \ref{models}) are implemented in PyTorch \citep{pytorch} using Huggingface \citep{wolf-etal-2020-transformers} and trained on a single NVIDIA RTX A6000 GPU (48GB). We use Adafactor \citep{adafactor} to train all our models with the same hyperparameters adapted from \citet{mosbach2021on}: 
\begin{itemize}
    \item Learning rate: $3e-5$
    \item Batch size: 16
    \item Epochs: 10
    \item Scheduler: Linear decay with $10\%$ warmup
\end{itemize}

\section{Statistical Significance Test}
\label{appendix:statTest}
To demonstrate the effectiveness of our three clause-level edit representations (Section \ref{results}),  we perform McNemar’s Test \citep{McNemar} to measure the statistical significance of their results in comparison to CodeT5-SQL+Token-Level.
For each significance test between two models, we use the median results among our three runs to calculate the comparison matrix.
Then, we compute the $p$-values using \texttt{statsmodels}.\footnote{\url{https://www.statsmodels.org/dev/generated/statsmodels.stats.contingency\_tables.mcnemar.html}}
When $p < 0.05$, we reject the null hypothesis.
In other words, we consider the accuracy improvement statistically significant when $p < 0.05$.

\section{Additional Results} 
\label{appendix:otherResults}

\begin{table*}[t]
\small
\centering
\begin{tabular}{lcccccccc}
\toprule
\multirow{2}{*}{\textbf{Models}} & \multirow{2}{*}{\textbf{Query}} & \multirow{2}{*}{\textbf{Edit}} & \multicolumn{2}{c}{\textbf{CodeT5}} & \multicolumn{2}{c}{\textbf{BRIDGEv2}} & \multicolumn{2}{c}{\textbf{SmBoP}} \\
\cmidrule(lr){4-5}\cmidrule(lr){6-7}\cmidrule(lr){8-9}
 & & & EM & EX & EM & EX & EM & EX\\
\midrule
\multirow{3}{*}{CoditT5} & SQL & Token-Level & 26.1 (0.4) & 28.6 (1.0) & 25.8 (0.3) & 27.2 (0.6) & 28.1 (0.9) & 30.7 (0.7) \\
 & SQL & Clause-Level & 28.6 (0.4) & 31.3 (0.5) & 28.4 (0.5) & 30.0 (0.2) & 30.2 (0.8) & 33.4 (0.8) \\
 & PyDict & Clause-Level & 28.9 (0.6) & 32.3 (0.8) & 28.0 (0.1) & 30.1 (0.2) & 27.6 (0.1) & 30.9 (0.4) \\
\midrule
\multirow{3}{*}{CodeT5} & SQL & Token-Level & 32.1 (1.1) & 34.1 (1.2) & 31.8 (0.4) & 34.5 (0.8) & 34.2 (0.1) & 37.6 (0.1) \\
 & SQL & Clause-Level & \underline{36.5} (0.6) & \textbf{38.6} (0.5) & \textbf{35.9} (0.4) & \textbf{39.3} (1.3) & \textbf{36.1} (0.6) & 38.8 (0.5)  \\
 & PyDict & Clause-Level & 35.6 (0.9) & 37.9 (0.3) & 32.9 (1.0) & 34.8 (0.8) & 33.0 (0.2) & 36.3 (0.3) \\
\midrule
CodeT5$^*$ & \multirow{2}{*}{PyDict} & \multirow{2}{*}{Program} & 35.7 (0.8) & 37.9 (0.3) & \underline{34.8} (0.8) & \underline{38.3} (0.7) & \underline{36.0} (0.3) & \textbf{40.2} (0.5) \\
CodeT5 &  &  & \textbf{36.7} (0.2) & \underline{38.5} (0.6) & 34.5 (0.1) & 37.1 (0.2) & 35.6 (0.8) & \underline{39.0} (0.1) \\
\bottomrule
\end{tabular}
\caption{\label{tab:dev}
Exact Set Match (EM) and Execution Match (EX) accuracy on our held-out development set (Section \ref{dataSyn}). The \textbf{best performances} are in bold and the \underline{second bests} are underlined. $^*$We fine-tune the model to generate edit programs only (without resulting queries) and use Python interpreter to execute the edit actions.}
\end{table*}

\begin{table*}[t]
\small
\centering
\begin{tabular}{lcccccccc}
\toprule
\multirow{2}{*}{\textbf{Models}} & \multirow{2}{*}{\textbf{Query}} & \multirow{2}{*}{\textbf{Edit}} & \multicolumn{2}{c}{\textbf{CodeT5}} & \multicolumn{2}{c}{\textbf{BRIDGEv2}} & \multicolumn{2}{c}{\textbf{SmBoP}} \\
\cmidrule(lr){4-5}\cmidrule(lr){6-7}\cmidrule(lr){8-9}
 & & & EM & EX & EM & EX & EM & EX\\
\midrule
No Edit & N/A & N/A & 62.7 (-) & 63.6 (-) & 70.1 (-) & 68.2 (-) & 74.6 (-) & 75.3 (-) \\
\midrule
CodeT5$^*$ & \multirow{2}{*}{PyDict} & \multirow{2}{*}{Program} & 69.2 (0.4) & 68.4 (0.2) & 72.5 (0.4) & 73.1 (0.2) & 77.3 (0.4) & 77.6 (0.6) \\
CodeT5 &  &  & 69.0 (0.2) & 68.2 (0.1) & 72.5 (0.3) & 73.0 (0.6) & 78.0 (0.3) & 78.5 (0.3) \\
\midrule
CodeT5$^{\dagger}$ & PyDict & Program & \textbf{73.0} (0.7) & \textbf{72.9} (0.8) & \textbf{76.6} (0.4) & \textbf{78.1} (0.2) & \textbf{80.0} (0.3) & \textbf{81.2} (0.6) \\
\bottomrule
\end{tabular}
\caption{\label{tab:interaction}
Exact Set Match (EM) and Execution Match (EX) accuracy on Spider development set. The \textbf{best performances} are in bold. $^*$We fine-tune the model to generate edit programs only (without resulting queries) and use Python interpreter to execute the edit actions. $^{\dagger}$We simulate user interactions using gold SQL queries to choose edit actions during beam search (size 3) and then execute the chosen actions to get the resulting SQL parse.
}
\end{table*}

\paragraph{Results on our development set.}
\rc{We report model performances on our held-out development set (Section \ref{dataSyn}) in Table \ref{tab:dev}. 
During training, we select the best model by evaluating its EX and EM accuracy on the development set (Section \ref{eval}) every 500 steps. 
Surprisingly, we find that CodeT5-SQL+Clause-Level sometimes achieves the best performance.
For BRIDGEv2, it obtains 35.9 EM accuracy and 39.3 EX accuracy, while CodeT5-PyDict+Program only obtains 34.5 EM accuracy and 37.1 EX accuracy.
A possible explanation is that in comparison to the test set, our development set has SQL structures and databases that are more similar to the training set, while the test set has unseen SQL structures and less similar databases.
It may also indicate that CodeT5-SQL+Clause-Level overfits the synthetic training data and fails to generalize to realistic test data.}

\paragraph{Results for simulated interaction experiments.}
\rc{To show the potential of using our model in an interactive framework, we extend our main experiments (Section \ref{results}) by adding simulated user interactions.
Since our model uses beam search to decode the edit actions $\mathbf{e} = \{e_1, e_2, ..., e_n\}$ and the resulting correct SQL query $\mathbf{q_{+}}$ (Equation \ref{eq:generation}), we simulate user interactions to select one edit action $e_i$ at a time from the beam results.}

\rc{At each time step $t$, we prompt the decoder with previously selected edit actions $e_1, ..., e_{t-1}$ to complete the sequence $e_{t}, ..., e_{n}, \mathbf{q_{+}}$ using beam search with size 3.
Then, we use gold SQL annotations to simulate the user interaction, which selects an edit action $e_{t}$ from the three candidates at step $t$ or chooses to skip the current step when all three candidates are wrong.
If skipping, the user continues to check the consequent edit actions $e_{t+j}$ ($j = 1, 2, ..., n - t$) until it selects the next edit action.
When the interaction finishes, we append the selected edit action to the prompt and let the model regenerate a completion with the new prompt for the next step's interaction.
Having simulated interactions for all edit actions, we do not use the generated $\mathbf{q_{+}}$ directly because some edit actions are skipped.
Instead, we execute the selected ones on the initial SQL query to derive the final query.}

\rc{As shown in Table \ref{tab:interaction}, when collaborating with a simulated user, our error correction model can further improve the base parsers' accuracy.
Compared to its performance without using any interactions, our model achieves up to 4.1 point more absolute improvement on EM accuracy ($72.5 \rightarrow 76.6$; BRIDGEv2) and 5.0 point more absolute improvement on EX accuracy ($73.1 \rightarrow 78.1$; BRIDGEv2).
With these results for simulated interaction experiments, we deem that incorporating our error correction model into an interactive framework is a promising future direction.}

\setcounter{table}{0}
\renewcommand\thetable{\Alph{section}.\arabic{table}}
\setcounter{figure}{0}
\renewcommand\thefigure{\Alph{section}.\arabic{figure}}

\newpage
\section{More Representation Examples}
\label{appendix:examples}
We provide two more examples in Table \ref{tab:rep2} and \ref{tab:rep3} to demonstrate how we represent SQL with subqueries and their edits (Section \ref{editRep}). We also show different representations for \textit{Insert} and \textit{Delete} edit actions. 

\begin{table*}
\small
\centering
\begin{tabularx}{\textwidth}{cXcX}
\toprule
\multicolumn{2}{c}{\textbf{Query Representation}} & \multicolumn{2}{c}{\textbf{Edit Representation}} \\
\cmidrule(lr){1-2}\cmidrule(lr){3-4}
\multirow{2}{*}{SQL} & \multirow{2}{\hsize}{\textcolor{gray}{select count(*) \\ from cars\_data \\ where cars\_data.accelerate > ( \\ \quad select \textcolor{red}{max(cars\_data.horsepower)} \\ \quad from cars\_data \\ )}} & Token-level & \textcolor{gray}{<ReplaceOld> \textcolor{red}{max(cars\_data.horsepower)} <ReplaceNew> \textcolor{red}{cars\_data.accelerate} <ReplaceEnd> <Insert> \textcolor{red}{order by cars\_data.horsepower desc limit 1} <InsertEnd>} \\
 & & Clause-level & \textcolor{gray}{<ReplaceOld> select \textcolor{red}{max(cars\_data.horsepower)} <ReplaceNew> select \textcolor{red}{cars\_data.accelerate} <ReplaceEnd> <Insert> \textcolor{red}{order by cars\_data.horsepower desc limit 1} <InsertEnd>} \\
\cmidrule(lr){1-2}\cmidrule(lr){3-4}
\multirow{2}{*}{PyDict} & \multirow{2}{\hsize}{\textcolor{gray}{sql = \{ \\ \quad "select": "select count(*)", \\ \quad  "from": "from cars\_data", \\ \quad "where": \{ \\ \quad \quad "clause": "where cars\_data.accelerate > (subquery0)", \\ \quad \quad  "subquery0": \{ \\ \quad \quad \quad "select": " select \textcolor{red}{max(cars\_data.horsepower)}", \\ \quad \quad \quad "from": "from cars\_data" \\ \quad \quad \} \\ \quad \} \\ \}}} & Clause-level & \textcolor{gray}{<ReplaceOld> "select": "select \textcolor{red}{max( cars\_data.horsepower)}" <ReplaceNew> "select": "select \textcolor{red}{cars\_data.accelerate}" <ReplaceEnd> <Insert> \textcolor{red}{"orderBy": "order by cars\_data.horsepower desc",  "limit": "limit 1"} <InsertEnd>} \newline\\
 &  & Program & \textcolor{gray}{sql["where"]["subquery0"]["select"] = "select \textcolor{red}{cars\_data.accelerate}" \newline \textcolor{red}{sql["where"]["subquery0"]["orderBy"] = "order by cars\_data.horsepower desc" \newline sql["where"]["subquery0"]["limit"] = "limit 1"}} \newline  \\
\bottomrule
\end{tabularx}
\caption{\label{tab:rep2}
Example representations for a wrong SQL query \textit{that contains a nested subquery} and its edit actions (including \textit{Insert} edits). The corresponding natural language utterance is ``What is the number of cars with a greater accelerate than the one with the most horsepower?''
}
\end{table*}

\begin{table*}
\small
\centering
\begin{tabularx}{\textwidth}{cXcX}
\toprule
\multicolumn{2}{c}{\textbf{Query Representation}} & \multicolumn{2}{c}{\textbf{Edit Representation}} \\
\cmidrule(lr){1-2}\cmidrule(lr){3-4}
\multirow{2}{*}{SQL} & \multirow{2}{\hsize}{\textcolor{gray}{select employee.name \\ from employee join evaluation on employee.employee\_id = evaluation.employee\_id \\ \textcolor{red}{group by evaluation.employee\_id"} \\ order by \textcolor{red}{sum(evaluation.bonus)} desc \\ limit 1}} & Token-level & \textcolor{gray}{<Delete> \textcolor{red}{group by evaluation.employee\_id} <DeleteEnd> <Delete> \textcolor{red}{sum(} <DeleteEnd> <Delete> \textcolor{red}{)} <DeleteEnd>} \\
 & & Clause-level & \textcolor{gray}{<Delete> \textcolor{red}{group by evaluation.employee\_id} <DeleteEnd> <ReplaceOld> order by \textcolor{red}{sum(evaluation.bonus)} desc <ReplaceNew> order by \textcolor{red}{evaluation.bonus} desc <ReplaceEnd>} \\
\cmidrule(lr){1-2}\cmidrule(lr){3-4}
\multirow{2}{*}{PyDict} & \multirow{2}{\hsize}{\textcolor{gray}{
sql = \{ \\ \quad "select": "select employee.name", \\ \quad "from": "from employee join evaluation on employee.employee\_id = evaluation.employee\_id", \\ \quad \textcolor{red}{"groupBy": "group by evaluation.employee\_id"}, \\ \quad "orderBy": "order by \textcolor{red}{sum(evaluation.bonus)} desc", \\ \quad "limit": "limit 1" \\ \}}} & Clause-level & \textcolor{gray}{<Delete> \textcolor{red}{"groupBy": "group by evaluation.employee\_id"} <DeleteEnd> <ReplaceOld> "orderBy": "order by \textcolor{red}{sum(evaluation.bonus)} desc" <ReplaceNew> "orderBy": "order by \textcolor{red}{evaluation.bonus} desc" <ReplaceEnd>} \newline\\
 &  & Program & \textcolor{gray}{\textcolor{red}{sql.pop("groupBy")}\newline sql["orderBy"] = "order by \textcolor{red}{evaluation.bonus} desc"} \newline\newline \\
\bottomrule
\end{tabularx}
\caption{\label{tab:rep3}
Example representations for a wrong SQL query and its edit actions (including \textit{Delete} edits). The corresponding natural language utterance is ``Find the name of the employee who got the highest one time bonus.''
}
\end{table*}

\end{document}